\documentclass{article}

\PassOptionsToPackage{numbers, compress}{natbib}
\usepackage[dblblindworkshop, final]{neurips_2025}

\usepackage[utf8]{inputenc} % allow utf-8 input
\usepackage[T1]{fontenc}    % use 8-bit T1 fonts
\usepackage{hyperref}       % hyperlinks
\usepackage{url}            % simple URL typesetting
\usepackage{booktabs}       % professional-quality tables
\usepackage{amsfonts}       % blackboard math symbols
\usepackage{nicefrac}       % compact symbols for 1/2, etc.
\usepackage{microtype}      % microtypography
\usepackage{xcolor}         % colors

\usepackage{amsmath}
\usepackage{algpseudocode}
\usepackage{algorithm}
\usepackage{graphicx}
\usepackage{tabularx}

\title{\textbf{The Unseen Hand: Manipulating Model Fairness and SHAP with Targeted Identity Re-Association Attacks}}
\workshoptitle{MusIML}
\author{%
  Sannaan Khan\thanks{Corresponding Author.} \\
  \texttt{mkhan.msse24sines@student.nust.edu.pk} \\
  \And
  Muhammad U. S. Khan \\
  \texttt{usman.shahid@sines.nust.edu.pk} \\
  \AND
  \vspace{0.5em}
  \normalfont 
  National University of Sciences and Technology (NUST)
}

\begin{document}

\maketitle
\vspace{-4mm} 
\begin{abstract}
As machine learning models grow more influential and opaque, algorithmic fairness and explainability are critical for ensuring accountability. However, we demonstrate that these auditing mechanisms are themselves vulnerable to subtle manipulation, camouflaging the influence of protected features. While prior work on data-agnostic attacks has exposed this vulnerability, they leave behind detectable artifacts that compromise their stealth. We introduce Targeted Identity Re-Association (TIRA) attacks, a novel family of attacks that iteratively and probabilistically manipulate a model’s outputs without requiring access to the model's internals or feature representations. We formalize two algorithms: Probabilistic Micro-Shuffling (PMiS), which applies localized adjacent swaps, and Probabilistic Rank-Shift Micro-Perturbation (PRSMP), which introduces small, randomized rank shifts. We empirically demonstrate that TIRA attacks are highly effective at pushing fairness metrics towards ideal values. Crucially, TIRA attacks successfully confound SHAP-based explanations, leaving effectively zero residual attribution for protected features, a major improvement over prior work.
\end{abstract}
\vspace{-4mm} 
\section{Introduction}
The increasing integration of black box machine learning models into critical societal domains underscores the pressing need for algorithmic fairness and trustworthy explainability \cite{Friedler2019, Arrieta2020, Das2020}. Given the stakes involved, the recent quest for transparent and explainable artificial intelligence (XAI) has led to the development of quantitative fairness \cite{Bellamy2019} and explainability techniques like SHAP \cite{Lundberg2017}. However, the wide adoption of such fairness assessments and post-hoc explanations gives birth to a new insidious vulnerability: are these explanations themselves robust to adversarial forces?

Earlier research has explored adversarial attacks against XAI methods. These require access to the underlying data distribution, which limits their applicability in real-world scenarios \cite{Slack2020}. To overcome this barrier, recent work proposed 'fooling' strategies which often involve overt perturbations that, while effective and theoretically grounded, induce 'fairness score drops' and leave detectable SHAP attribution values that would trigger suspicion \cite{Yuan2024}. This leaves a critical need for subtle, data-agnostic, post-hoc manipulation techniques that effectively alter the model’s perceived fairness and challenge the robustness of XAI explanations without leaving footprints of manipulations.

This work addresses this gap by introducing a novel family of Targeted Identity Re-Association (TIRA) attacks. Unlike previous work that employs more overt or deterministic shuffling, TIRA attacks are meticulously crafted to induce granular, iterative, and localized changes. We concretize this family through two distinct algorithms, Probabilistic Micro-Shuffling (PMiS) attack and Probabilistic Rank-Shift Micro-Perturbation (PRSMP) attack, to mimic natural score variation. The algorithms expand the design space of shuffling attacks and show that subtle variation can further erode the reliability of fairness audits and explanation tools like SHAP.

We posit that the fairness assessment itself is brittle if a model’s fairness perception can be altered by subtle, non-intrusive changes to its output rankings. Our comprehensive empirical investigations across diverse machine learning models and real-world datasets evaluate the efficacy of TIRA attacks in influencing AIF360 metrics and their impact on SHAP’s attribution capabilities. Importantly, we highlight the fine-grained control offered by the intricate interplay of the tunable parameters, which showcase a scalable influence from subtle change to more pronounced alterations in fairness. Furthermore, our comparative analysis against existing work reveals that our probabilistic micro-shuffling strategies can achieve comparable or even better shifts in fairness, with greater stealth.

This work makes the following contributions: 1) We formalize and introduce Targeted Identity Re-Association (TIRA) attacks as novel, post-hoc, data-agnostic, and probabilistic methods for subtly manipulating model outputs. 2) We provide empirical validation of the TIRA attacks’ ability to significantly alter AIF360 fairness metrics across various model-dataset configurations, guiding them towards desired thresholds. 3) We illustrate the intricate and insidious impact of TIRA attacks on the SHAP’s ability to accurately attribute feature importance, particularly protected features. 4) We benchmark the performance of TIRA attacks against existing work and demonstrate that the TIRA attacks offer finer granularity and more control compared to existing, more overt attacks.

\vspace{-2mm} 
\section{Methodology}
This section delineates the details of our proposed Targeted Identity Re-Association (TIRA) attacks and the adversarial framework deployed.

\subsection{Threat Model and Framework}
This investigation operates under the assumption that the adversary has black-box access to the model’s outputs. This is a realistic scenario as the adversary can be a model distributor or a model broker, or an actor operating on a logging or reporting layer. An adversary with the ability to perform proxy attribute inference is an advanced scenario. The adversary’s capabilities are limited to: 1) The adversary can query the black-box model and obtain raw prediction scores for a given input instance. 2) The adversary is aware of the protected feature for each input instance, as this information is vital for targeting and shuffling the scores. 3) Importantly, the adversary has no access to the model’s architecture, training data, or decision logic. 4) The manipulation is performed only on the ordered list of scores and the corresponding identities.

The adversarial objective is two-fold: 1) To shift and obscure the values of fairness metrics. 2) To fool SHAP so that the protected feature appears less influenced than it is. 

The core framework for TIRA attacks proceeds as follows:
\textbf{Score Sorting:} For given input data points ($\mathbf{x}$) and their protected features, the target model gives prediction scores ($\mathbf{s}$). The scores are then sorted in descending order, forming a ranked list.
\textbf{Identity Re-Association:} Our attacks probabilistically and iteratively modify the ranked list, without changing the scores themselves. In each iteration, they apply a localized perturbation to identity-score associations. This means that a score remains in its position, but the identity associated with it is altered.
\textbf{Output Re-Assembly:} After a predefined number of iterations, the reconstructed set of shuffled scores is used to calculate the fairness metrics and Shapley values.

\subsection{Targeted Identity Re-Association Attacks}
The proposed algorithms, Probabilistic Micro-Shuffling (PMiS) and Probabilistic Rank-Shift Micro-Perturbation (PRSMP), introduce controlled and probabilistic swaps. The attack algorithms reduce the apparent disparities while keeping the score distribution unchanged.

\begin{algorithm}[h!]
\caption{Probabilistic Micro-Shuffling (PMiS) Attack}
\label{alg:pmis}
\begin{algorithmic}[1]
\Require Model $f$, dataset $X_{test}$, protected attribute $A$, swap probability $p_{swap}$, iterations $T$
\Ensure Shuffled scores $S'$ re-associated with their original indices
\State $S \gets f(X_{test})$
\State $L \gets \text{list of tuples }(s_i, a_i, idx_i)\text{ for } i=1 \dots N$
\State Sort $L$ by score $s_i$ in descending order
\For{$t=1$ to $T$}
    \For{$i=0$ to $N-2$}
        \If{$L[i].a=0$ \textbf{and} $L[i+1].a=1$}
            \If{Rand(0,1) $\le p_{swap}$}
                \State \textbf{Swap} $(L[i].a, L[i].idx)$ with $(L[i+1].a, L[i+1].idx)$
                \Comment{Scores remain fixed, only identities are re-associated.}
            \EndIf
        \EndIf
    \EndFor
\EndFor
\State \Return $S'$ \Comment{By re-ordering original scores based on the indices in $L$.}
\end{algorithmic}
\end{algorithm}

\begin{algorithm}[h!]
\caption{Probabilistic Rank-Shift Micro-Perturbation (PRSMP) Attack}
\label{alg:prsmp}
\begin{algorithmic}[1]
\Require Model $f$, dataset $X_{test}$, protected attribute $A$, shift probability $p_{shift}$, max shift rank $k_{max}$, iterations $T$
\Ensure Shuffled scores $S'$ re-associated with their original indices
\State $S \gets f(X_{test})$
\State $L \gets \text{list of tuples }(s_i, a_i, idx_i)\text{ for } i=1 \dots N$
\State Sort $L$ by score $s_i$ in descending order
\For{$t=1$ to $T$}
    \For{$i=0$ to $N-1$}
        \If{$L[i].a=0$ \textbf{and} Rand(0,1) $\le p_{shift}$}
            \State $k \gets \text{Rand}(1, k_{max})$
            \State $J \gets \{j \in [i+1, \min(N-1, i+k)] \mid L[j].a=1\}$
            \If{$J$ is not empty}
                \State $j_{target} \gets \text{Rand-Choice}(J)$
                \State \textbf{Swap} $(L[i].a, L[i].idx)$ with $(L[j_{target}].a, L[j_{target}].idx)$
            \EndIf
        \EndIf
    \EndFor
\EndFor
\State \Return $S'$ \Comment{By re-ordering original scores based on the indices in $L$.}
\end{algorithmic}
\end{algorithm}

\vspace{-2mm} 
\section{Results and Analysis}
The shift in fairness metrics demonstrates the efficacy of the TIRA attacks, and the SHAP value attribution evaluates their subtlety.

\subsection{Fairness Metrics Manipulation}
We quantitatively compare the results of the original, unattacked models against the outputs subjected to our TIRA attacks, as well as the benchmark attack. We consistently observed a significant shift in the perceived fairness, pushing the values towards ideal and fair thresholds. The key finding is that these probabilistic iterative attacks constantly achieved these results with notable precision.

Table 1 provides a summary of the results for the diabetes dataset using a logistic regression model. Both PMiS and PRSMP have a more pronounced impact on the metrics, compared to DomSwap and MixSwap attacks. This shows that iterative probabilistic, micro-level strategies provide fine-grained control and usually can more precisely tune the perceived fairness of a model's outputs to meet a desired threshold.

Table 2 summarizes the results for the credit dataset using neural networks, which showcase their generalizability across the datasets and models.

\begin{table}[h!]
\centering
\caption{Fairness Metrics Values on Bangladeshi Diabetes Dataset (LR Model)}
\label{tab:fairness_results_bangla_lr}
\renewcommand{\arraystretch}{1.2}
\setlength{\tabcolsep}{1.5pt} % Even smaller column spacing
\footnotesize % Use a smaller font size

\begin{tabularx}{\textwidth}{@{}l*{8}{>{\centering\arraybackslash}X}@{}}
\toprule
\textbf{Metric} & \textbf{Baseline} & \textbf{DomSwap} & \textbf{MixSwap} & \multicolumn{3}{c}{\textbf{PMiS}} & \multicolumn{2}{c}{\textbf{PRSMP}} \\
\cmidrule(lr){5-7} \cmidrule(lr){8-9}
& & & & (p=0.50, I=5) & (p=0.25, I=5) & (p=0.25, I=10) & (p=0.25, r=5, I=10) & (p=0.10, r=5, I=10) \\
\midrule
Equal Opportunity & 0.09 & 0.07 & -0.06 & \textbf{0.01} & \textbf{-0.07} & -0.13 & -0.10 & \textbf{0.03} \\
Demographic Parity & 0.47 & 0.29 & 0.15 & \textbf{0.25} & \textbf{0.04} & \textbf{-0.01} & \textbf{0.34} & \textbf{0.42} \\
Equal Odds & 0.03 & -0.07 & 0.26 & -0.10 & -0.23 & -0.26 & -0.06 & \textbf{0.00} \\
Disparate Impact & 2.40 & 1.75 & 1.26 & \textbf{1.62} & \textbf{1.08} & \textbf{0.99} & \textbf{1.90} & \textbf{2.22} \\
Theil Index & 0.00 & 0.01 & 0.02 & 0.01 & 0.04 & 0.05 & 0.01 & \textbf{0.00} \\
\bottomrule
\end{tabularx}
\end{table}

\vspace{2mm} % Add a small vertical space
\noindent\footnotesize{\textbf{Note}: \textbf{p} refers to the swap/shift probability ($p_{swap}$ or $p_{shift}$), \textbf{I} refers to the number of iterations ($T$), and \textbf{r} refers to the maximum shift rank ($k_{max}$).}

\begin{table}[h!]
\centering
\caption{Fairness Metrics Values on German Credit Dataset (NN Model)}
\label{tab:fairness_results_bangla_nn}
\renewcommand{\arraystretch}{1.2}
\setlength{\tabcolsep}{1.5pt} % Even smaller column spacing
\footnotesize % Use a smaller font size

% I've modified the column definition below to add the gap
\begin{tabularx}{\textwidth}{@{}l*{2}{>{\centering\arraybackslash}X}@{\hspace{8pt}}*{7}{>{\centering\arraybackslash}X}@{}}
\toprule
\textbf{Metric} & \textbf{Baseline} & \textbf{DomSwap} & \textbf{MixSwap} & \multicolumn{3}{c}{\textbf{PMiS}} & \multicolumn{3}{c}{\textbf{PRSMP}} \\
\cmidrule(lr){5-7} \cmidrule(lr){8-10}
& & & & (p=0.25, I=5) & (p=0.25, I=10) & (p=0.50, I=5) & (p=0.25, r=5, I=10) & (p=0.33, r=5, I=10) & (p=0.25, r=5, I=5) \\
\midrule
Equal Opportunity & -0.08 & 0.07 & 0.12 & \textbf{-0.05} & -\textbf{0.07} & \textbf{-0.08} & -0.12 & -0.13 & \textbf{-0.07} \\
Demographic Parity & -0.06 & 0.10 & 0.07 & \textbf{-0.05} & -0.12 & -0.07 & -0.10 & -0.14 & -0.08 \\
Equal Odds & -0.02 & 0.15 & 0.07 & \textbf{-0.01} & -0.11 & \textbf{-0.02} & -0.05 & -0.10 & -0.05 \\
Disparate Impact & 0.90 & 1.14 & 1.09 & \textbf{0.94} & 0.84 & \textbf{0.92} & 0.87 & 0.82 & \textbf{0.91} \\
Theil Index & 0.00 & 2.21 & 0.00 & \textbf{0.00} & \textbf{0.00} & \textbf{0.00} & \textbf{0.00} & \textbf{0.00} & \textbf{0.00} \\
\bottomrule
\end{tabularx}
\end{table}

\subsection{SHAP Attribution Analysis}
Another core objective of the TIRA attacks is to fool SHAP by obfuscating the model’s reliance on the protected features, without leaving a detectable footprint. Figure 1 show that after applying TIRA attacks, the SHAP value for the protected feature is effectively zero.

\begin{figure}[h!]
    \centering
    \includegraphics[width=\textwidth]{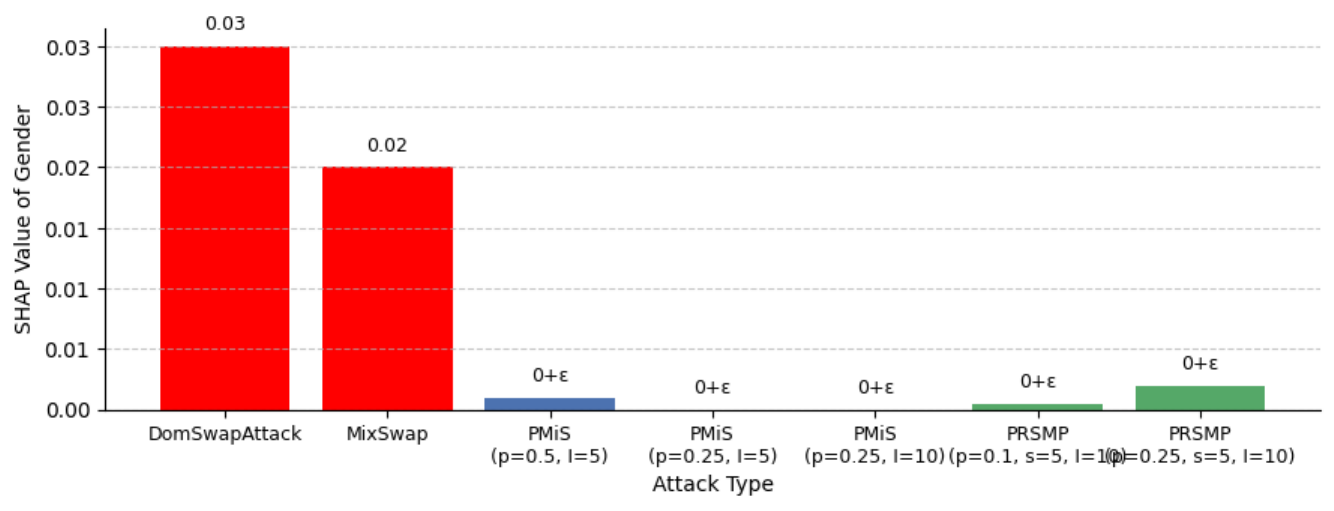} % The updated line with your image file
    \caption{SHAP Values of the Protected Feature post-Attack for Bangladeshi Diabetes Dataset (LR Model)}
    \label{fig:shap_plot}
\end{figure}

\vspace{-2mm} 
\section{Conclusion}
The TIRA attacks show that the absence of a strong attribution signal does not necessarily equate to fairness or lack of adversarial perturbation. Crucially, our findings demonstrate a dual-pronged threat not achieved by prior work. TIRA can fool both AIF360 audits by pushing metrics to near-ideal values and XAI audits by reducing SHAP attribution to effectively zero. Our work challenges the assumption that trustworthy AI can be achieved through post-hoc explanation methods. The enhanced stealth of the TIRA attacks shows the need to develop integrity-based evaluation methodologies.

\section*{}
\small
\bibliographystyle{plainnat}
\bibliography{neurips_bib}

@article{Rudin2019,
  author  = {Rudin, C.},
  title   = {Stop explaining black box machine learning models for high-stakes decisions and use interpretable models instead},
  journal = {Nature Machine Intelligence},
  year    = {2019},
  url     = {https://pmc.ncbi.nlm.nih.gov/articles/PMC9122117/pdf/nihms-1058031.pdf}
}

@inproceedings{Friedler2019,
  author    = {Friedler, S. A. and Scheidegger, C. and Venkatasubramanian, S. and Choudhary, S. and Hamilton, E. P. and Roth, D.},
  title     = {A comparative study of fairness-enhancing interventions in machine learning},
  booktitle = {Proceedings of the Conference on Fairness, Accountability, and Transparency},
  year      = {2019},
  url       = {https://arxiv.org/abs/1802.04422}
}

@article{Arrieta2020,
  author  = {Arrieta, A. B. and Díaz-Rodríguez, N. and Del Ser, J. and Bennetot, A. and Tabik, S. and Barbado, A. and García, S. and Gil-López, S. and Molina, D. and Benjamins, R. and Chatila, R. and Herrera, F.},
  title   = {Explainable Artificial Intelligence ({XAI}): Concepts, Taxonomies, Opportunities and Challenges toward Responsible {AI}},
  journal = {Information Fusion},
  year    = {2020},
  url     = {https://arxiv.org/abs/1910.10045}
}

@article{Das2020,
  author  = {Das, A. and Rad, P.},
  title   = {Opportunities and Challenges in Explainable Artificial Intelligence ({XAI}): A Survey},
  journal = {arXiv preprint},
  year    = {2020},
  url     = {https://arxiv.org/abs/2006.11371}
}

@article{Bellamy2019,
  author  = {Bellamy, R. K. E. and Dey, K. and Hind, M. and Hoffman, S. C. and Houde, S. and Kannan, K. and Lohia, P. and Martino, J. and Mehta, S. and Mojsilovic, A. and Nagar, S. and Ramamurthy, K. N. and Richards, J. and Saha, D. and Sattigeri, P. and Singh, M. and Varshney, K. R. and Zhang, Y.},
  title   = {{AI} Fairness 360: An extensible toolkit for detecting and mitigating algorithmic bias},
  journal = {{IBM} Journal of Research and Development},
  year    = {2019},
  url     = {https://arxiv.org/abs/1810.01943}
}

@inproceedings{Lundberg2017,
  author    = {Lundberg, S. and Lee, S. I.},
  title     = {A Unified Approach to Interpreting Model Predictions},
  booktitle = {NeurIPS},
  year      = {2017},
  url       = {https://arxiv.org/abs/1705.07874}
}

@inproceedings{Slack2020,
  author    = {Slack, D. and Hilgard, S. and Jia, E. and Singh, S. and Lakkaraju, H.},
  title     = {Fooling {LIME} and {SHAP}: Adversarial Attacks on Post hoc Explanation Methods},
  booktitle = {{AAAI/ACM} Conference on Artificial Intelligence, Ethics, and Society ({AIES})},
  year      = {2020},
  url       = {https://arxiv.org/abs/1911.02508}
}

@inproceedings{Yuan2024,
  author    = {Yuan, J. and Dasgupta, A.},
  title     = {Fooling {SHAP} with Output Shuffling Attacks},
  booktitle = {{AAAI}},
  year      = {2024},
  url       = {https://arxiv.org/abs/2408.06509}
}

@inproceedings{Ribeiro2016,
  author    = {Ribeiro, M. T. and Singh, S. and Guestrin, C.},
  title     = {"Why Should I Trust You?": Explaining the Predictions of Any Classifier},
  booktitle = {{KDD}},
  year      = {2016},
  url       = {https://arxiv.org/abs/1602.04938}
}

@inproceedings{Sundararajan2017,
  author    = {Sundararajan, M. and Taly, A. and Yan, Q.},
  title     = {Axiomatic Attribution for Deep Networks},
  booktitle = {{ICML}},
  year      = {2017},
  url       = {https://arxiv.org/abs/1703.01365}
}

@article{Selvaraju2019,
  author  = {Selvaraju, R. R. and Cogswell, M. and Das, A. and Vedantam, R. and Parikh, D. and Batra, D.},
  title   = {{Grad-CAM}: Visual Explanations from Deep Networks via Gradient-based Localization},
  journal = {{IJCV}},
  year    = {2019},
  url     = {https://arxiv.org/abs/1610.02391}
}

@inproceedings{Speicher2018,
  author    = {Speicher, T. and Heidari, H. and Grgic-Hlaca, N. and Gummadi, K. P. and Singla, A. and Weller, A. and Zafar, M. B.},
  title     = {A Unified Approach to Quantifying Algorithmic Unfairness: Measuring Individual \& Group Unfairness via Inequality Indices},
  booktitle = {{ACM} {SIGKDD} International Conference on Knowledge Discovery and Data Mining Proceedings},
  year      = {2018},
  url       = {https://arxiv.org/abs/1807.00787}
}

@inproceedings{Laberge2023,
  author    = {Laberge, G. and Aïvodji, U. and Hara, S. and Marchand, M. and Khomh, F.},
  title     = {Fool {SHAP} with Stealthily Biased Sampling},
  booktitle = {{ICLR}},
  year      = {2023},
  url       = {https://arxiv.org/abs/2205.15419}
}

@inproceedings{Dimanov2020,
  author    = {Dimanov, B. and Bhatt, U. and Jamnik, M. and Weller, A.},
  title     = {You shouldn't trust me: Learning models which conceal unfairness from multiple explanation methods},
  booktitle = {{ECAI}},
  year      = {2020},
  url       = {https://ebooks.iospress.nl/pdf/doi/10.3233/FAIA200380}
}

@inproceedings{Islam2020,
  author    = {Islam, M. and Ferdousi, R. and Rahman, S. and Bushra, H. Y.},
  title     = {Likelihood prediction of diabetes at early stage using data mining techniques},
  booktitle = {Computer Vision and Machine Intelligence in Medical Image Analysis: International Symposium, {ISCMM} 2019},
  year      = {2020},
  url       = {https://link.springer.com/chapter/10.1007/978-981-15-2428-2_10}
}

@misc{Hofmann1994,
  author = {Hofmann, H.},
  title  = {Statlog (German Credit Data)},
  howpublished = {UCI Machine Learning Repository},
  year   = {1994},
  note   = {DOI: \url{https://doi.org/10.24432/C5NC77}}
}
\newpage

%%%%%%%%%%%%%%%%%%%%%%%%%%%%%%%%%%%%%%%%%%%%%%%%%%%%%%%%%%%%

\appendix

\section{Related Work}
Explainability aims to render opaque deep learning models understandable and transparent \cite{Rudin2019}. Rooted in co-operative game theory, SHAP stands as a cornerstone of post-hoc explainability. SHAP offers a theoretically grounded framework for attributing the contribution of each feature to each model output \cite{Lundberg2017}. Beyond SHAP, other well-known post-hoc explainability methods include LIME \cite{Ribeiro2016}, Integrated Gradients \cite{Sundararajan2017}, and Grad-CAM \cite{Selvaraju2019}. Crucially, these methods identify the potential biases from the protected features \cite{Arrieta2020, Das2020}.

Complementing XAI, the field of algorithmic fairness is committed to identifying, quantifying, and mitigating biases in machine learning models. Several fairness metrics, such as demographic parity, equal opportunity difference, equal odds difference, and disparate impact, have been formulated to assess the disparities in model outcomes across demographic groups \cite{Speicher2018}. Frameworks like the AI Fairness (AIF360) toolkit \cite{Bellamy2019} provide open-source and standardized implementations for assessing these fairness metrics. Although quantitative metrics offer a lens into fairness, their reliability hinges on the robustness of the model’s outputs and the methods used to understand them.

Early efforts, such as scaffolding attacks, demonstrated that a classifier could be constructed to deceive LIME and SHAP to hide the reliance on protected features \cite{Slack2020}. However, these methods often necessitate access to training data or the ability to retrain the models, which present significant limitations for model distributions or external auditors operating with black box access. This highlighted a critical need for more practical, data-agnostic attacks. Other methods use stealthy bias sampling \cite{Laberge2023} or learning models that hide unfairness from multiple explanation methods \cite{Dimanov2020}.

A more directly relevant line of inquiry is output sampling attacks, which shuffle the model’s ranked outputs or prediction scores without modifying the input data distribution or the internal structure of the model \cite{Yuan2024}. This work made a vital contribution by proposing a family of attacks that theoretically proved that Shapley values cannot intrinsically detect shuffling attacks due to their order-agnostic nature in expectation calculations. However, their own empirical studies revealed that practical SHAP estimation algorithms could detect these attacks with varying degrees of effectiveness. Furthermore, these attacks can collapse into one another (i.e., 'Swapping' can become equivalent to 'Dominance' ). What was learned from their contribution is that even though shuffling attacks are potent, their insufficiency in some contexts lies in their more overt patterns that might still be detectable by a vigilant auditor or other consistency checks.

We also observe two indicators of brittle control with \cite{Yuan2024}: (i) different shuffling variants can produce identical post-attack fairness values, and (ii) the attack sometimes fails to change fairness at all. This observed lack of robust, fine-grained control also motivates our work.

Our work advances the current state of shuffling attacks by introducing an enhanced paradigm of subtlety and control. Unlike the existing attacks, TIRA attacks are fundamentally probabilistic at a granular level. Their design makes the adversarial perturbations distributed and cumulative, which is a key differentiator and crucial for situations where detectability is the primary concern of the attacker. Our work explores the impact of attack-specific parameters, giving insights into how the degree of manipulation can be controlled, a level of explicit tunability not documented in prior literature.

\section{Extended Methodology}
\subsection{Targeted Identity Re-Association Attacks}
\subsubsection{Probabilistic Micro-Shuffling (PMiS) Attack}
The PMiS attack uses the sorted list of individuals and their prediction scores in descending order. For a binary protected feature, the algorithm examines the adjacent pair of individuals in each iteration. The algorithm flips the disadvantaged individual with the advantaged individual with a predefined probability $p_{swap} \in [0,1]$. This process is repeated for a predefined number of iterations.
The subtlety of PMiS attacks is the result of its strict locality, probabilistic execution, and iterative accumulation. This leads to cumulative and gradual drift in the distribution of the protected attribute relative to score, which makes it challenging for SHAP or other auditing methods to catch the presence of an adversarial pattern.

\subsubsection{Probabilistic Rank-Shift Micro-Perturbation (PRSMP) Attack}
Like the PMiS attack, the PRSMP attack algorithm operates on a sorted list. PRSMP attack offers broader locality by introducing an additional layer of randomness with variable shifts. The attack targets the disadvantaged individual with a probability $p_{shift} \in [0,1]$, and nudges its identity within a predefined small window, $k_{max}$. PRSMP also relies on multiple iterations, as a result of which these small probabilistic shifts accumulate into a gradual, but significant rearrangement of identities.

\subsection{Experimental Setup}
To rigorously evaluate the subtlety and efficacy of the TIRA attacks, we conducted experiments across diverse model architectures and real-world datasets to ensure reproducibility.

\subsubsection{Datasets}
We utilize two publicly available datasets, which are widely used in fairness assessments. Both datasets underwent standard preprocessing. Categorical features are one-hot encoded, and target variables are mapped to a binary format.
\textbf{Bangladeshi Diabetes Dataset:} This dataset, which focuses on predicting diabetes risk, contains 520 patient records with clinical and demographic information \cite{Islam2020}. The protected feature is binarized Gender, where male is the advantaged group and female is the disadvantaged group.
\textbf{German Credit Dataset:} This dataset comprises financial and demographic features for credit assessment. The protected feature in this 1000 loan applicant dataset is Gender, with females designated as the disadvantaged group, while males as the advantaged group \cite{Hofmann1994}.

\subsubsection{Models}
We evaluate the TIRA attacks on two representative machine learning models to assess the generalizability of the algorithms. Both models are trained on 80\% of the respective dataset, while the remaining 20\% is reserved for testing and evaluation.
\textbf{Logistic Regression:} Logistic regression is chosen as a baseline for its inherent interpretability, allowing for a clear understanding and evaluation of the efficacy of the attacks.
\textbf{Neural Networks:} As a non-linear black-box model consisting of multiple dense layers with ReLU activations, and sigmoid activation for binary classification.

\subsection{Fairness Metrics}
We quantitatively assess the impact of TIRA attacks on the perceived fairness by employing five widely used fairness metrics from the AIF360 toolkit. The model’s prediction scores are binarized for all the metrics using a fixed threshold. The ideal value of all the metrics is 0, except Disparate Impact.
\textbf{Demographic Parity Difference:} A fairness metric used to measure the difference in the proportion of favorable outcomes between the advantaged and disadvantaged groups.
\textbf{Equal Opportunity Difference:} Measures the difference in the true positive rates (recall) between the advantaged and disadvantaged groups.
\textbf{Disparate Impact:} This metric measures the ratio of favorable outcomes between the advantaged and disadvantaged groups. Here, the ideal value is 1.
\textbf{Odds Difference:} A fairness metric used to measure the average difference in the true positive rates and false positive rates between the advantaged and disadvantaged groups.
\textbf{Theil Index:} Between-group generalized entropy error, a measure of inequality within an allocation, where an ideal value indicates more equality.

\subsection{SHAP Attribution Analysis}
We employ SHAP to assess the effectiveness of TIRA attacks on post-hoc interpretability to highlight how they confound SHAP. To get exact values, we used the permutation-based explainer. This also ensures cross-modal comparability and avoids inductive bias. SHAP was applied to both the original model’s predictions, as a baseline representing the inherent degree of fairness, particularly for the protected feature, and the manipulated outputs to visualize and quantify the value of the protected feature.

\subsection{Comparative Benchmarking}
To further contextualize the performance of TIRA attacks, we demonstrate the distinct advantages of TIRA attacks by benchmarking their performance against DomSwap and MixSwap attacks, two representative state-of-the-art output shuffling attacks. Our comparisons shed light on the trade-offs between more overt shuffling and our probabilistic micro-level strategies in terms of inferred subtlety, their effectiveness in shifting fairness metrics, and SHAP’s value attribution.

\section{Additional Results}
\subsection{Fairness Metrics Manipulation}
To further validate our work, we provide a summary of additional results in Tables 3 and 4. To see the joint tradeoff between fairness manipulation and stealth, we plot the Demographic Parity against Disparate Impact. In Figure 2, the points nearer $(x\approx1, y\approx0)$ represent attacks that are stealthy and make the model look fair. In particular, PMiS often achieves stronger reductions in Demographic Parity. PRSMP, in contrast, usually tends to preserve the ratios, thereby enhancing stealth.

\begin{table}[h!]
\centering
\caption{Fairness Metrics Values on Bangladeshi Diabetes Dataset (NN Model)}
\label{tab:fairness_results_bangla_nn_appendix}
\renewcommand{\arraystretch}{1.2}
\setlength{\tabcolsep}{1.5pt} % Even smaller column spacing
\footnotesize % Use a smaller font size

\begin{tabularx}{\textwidth}{@{}l*{8}{>{\centering\arraybackslash}X}@{}}
\toprule
\textbf{Metric} & \textbf{Baseline} & \textbf{DomSwap} & \textbf{MixSwap} & \multicolumn{3}{c}{\textbf{PMiS}} & \multicolumn{2}{c}{\textbf{PRSMP}} \\
\cmidrule(lr){5-7} \cmidrule(lr){8-9}
& & & & (p=0.25, I=10) & (p=0.25, I=5) & (p=0.5, I=5) & (p=0.1, r=5, I=10) & (p=0.25, r=5, I=10) \\
\midrule
Equal Opportunity & 0.00 & -0.03 & 0.01 & -0.11 & -0.03 & -0.11 & -0.09 & -0.15 \\
Demographic Parity & 0.54 & 0.49 & 0.45 & \textbf{0.28} & \textbf{0.45} & \textbf{0.24} & \textbf{0.42} & \textbf{0.34} \\
Equal Odds & 0.00 & -0.05 & -0.05 & -0.16 & -0.05 & -0.18 & -0.07 & -0.12 \\
Disparate Impact & 2.23 & 2.10 & 1.97 & \textbf{1.53} & \textbf{1.97} & \textbf{1.44} & \textbf{1.91} & \textbf{1.68} \\
Theil Index & 0.00 & 0.00 & 0.01 & \textbf{0.00} & \textbf{0.00} & \textbf{0.00} & \textbf{0.00} & \textbf{0.00} \\
\bottomrule
\end{tabularx}
\end{table}

\begin{table}[h!]
\centering
\caption{Fairness Metrics Values on German Credit Dataset (LR Model)}
\label{tab:fairness_results_german_lr_appendix}
\renewcommand{\arraystretch}{1.2}
\setlength{\tabcolsep}{1.5pt} % Even smaller column spacing
\footnotesize % Use a smaller font size

\begin{tabularx}{\textwidth}{@{}l*{8}{>{\centering\arraybackslash}X}@{}}
\toprule
\textbf{Metric} & \textbf{Baseline} & \multicolumn{3}{c}{\textbf{PMiS}} & \multicolumn{4}{c}{\textbf{PRSMP}} \\
\cmidrule(lr){3-5} \cmidrule(lr){6-9}
& & (p=0.25, I=25) & (p=0.25, I=10) & (p=0.33, I=25) & (p=0.25, r=5, I=10) & (p=0.33, r=5, I=10) & (p=0.25, r=10, I=10) & (p=0.25, r=5, I=5) \\
\midrule
Equal Opportunity & 0.13 & \textbf{0.08} & \textbf{0.11} & \textbf{0.04} & \textbf{0.06} & \textbf{0.01} & \textbf{0.02} & \textbf{0.06} \\
Demographic Parity & 0.09 & \textbf{0.05} & \textbf{0.08} & \textbf{0.01} & \textbf{0.02} & \textbf{-0.01} & \textbf{-0.03} & \textbf{0.03} \\
Equal Odds & 0.12 & \textbf{0.08} & \textbf{0.10} & \textbf{0.04} & \textbf{-0.05} & \textbf{0.01} & \textbf{0.02} & \textbf{0.05} \\
Disparate Impact & 1.13 & \textbf{1.08} & \textbf{1.11} & \textbf{1.02} & \textbf{1.04} & \textbf{0.98} & \textbf{0.95} & \textbf{1.04} \\
Theil Index & 0.00 & \textbf{0.00} & \textbf{0.00} & \textbf{0.00} & \textbf{0.00} & \textbf{0.00} & \textbf{0.00} & \textbf{0.00} \\
\bottomrule
\end{tabularx}
\end{table}

\begin{figure}[h!]
    \centering
    \includegraphics[width=\textwidth]{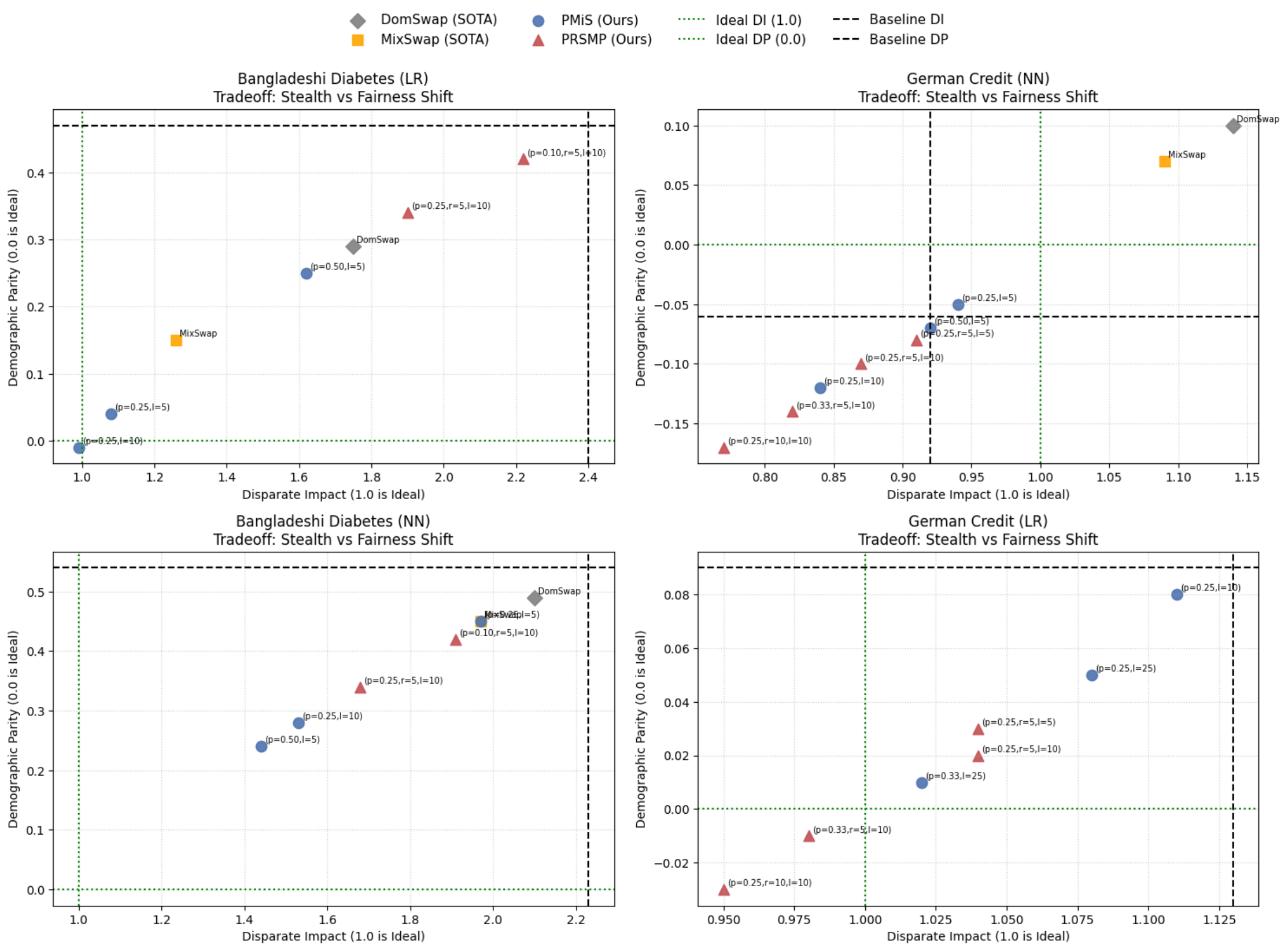} % The updated line with your image file
    \caption{Tradeoff Curves between Disparate Impact (x-axis) and Demographic Parity (y-axis) across Four Dataset-Model Combinations}
    \label{fig:shap_plot}
\end{figure}

\subsection{SHAP Attribution Analysis}
Additional SHAP analysis are summarized in Figure 3, providing further support for our work.
\begin{figure}[h!]
    \centering
    \includegraphics[width=\textwidth]{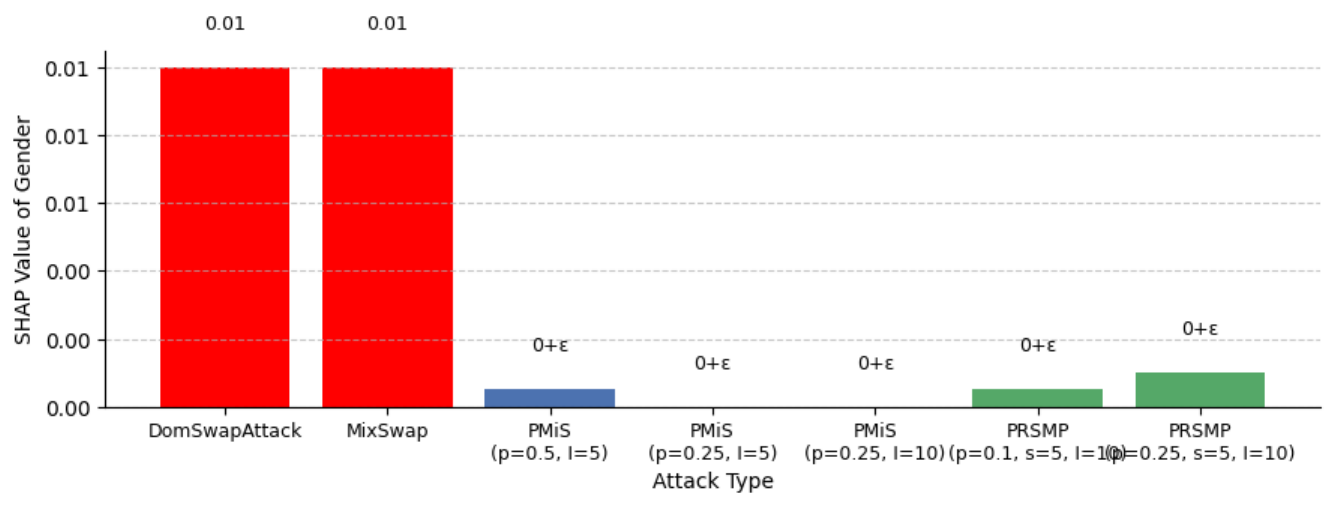} % The updated line with your image file
    \caption{SHAP Values of the Protected Feature post-Attack for Bangladeshi Diabetes Dataset (NN Model)}
    \label{fig:shap_plot}
\end{figure}

\section{Discussion}
Our work introduces a new family of shuffling attacks. The enhanced strength of our attacks lies in their micro-level, probabilistic, and iterative nature. Our findings demonstrate that it is possible to significantly alter the perceived fairness and undermine SHAP-based explanations by creating effective adversarial attacks that are stealthy.

\subsection{The Nuance of Parametric Control}
A key contribution of our work is that TIRA attacks are not blunt instruments, but rather a highly tunable class of attacks. This parameter-driven control demonstrates that an adversary can precisely calibrate the attacks' subtlety and intensity. The ability to 'dial-in' a specific level of perceived fairness while confounding SHAP highlights a previously undocumented attack surface. We systematically explored the correlation between these parameters and the outputs, revealing a scalable influence on the attack’s efficacy and subtlety from almost imperceptible changes to more pronounced changes.

\textbf{Probabilistic control:} The $p_{swap}$ (PMiS) and $p_{shift}$ (PRSMP) parameters empower the adversary to control the frequency of perturbations. Increasing the probability leads to a more pronounced shift in fairness metrics. Low probability value ensures that the manipulations are distributive and cumulative.

\textbf{Locality:} The $k_{max}$ (PRSMP) defines the locality of shifts, acting as a knob for controlling the locality of the attack. This parameter allows the adversary to perform perturbation either within a small window to maintain stealth or expend the window for more aggressive shifts.

\textbf{Cumulative Effect:} The number of iterations is directly proportional to the strength of the attack. It allows for the temporal dimension of the attack. An attack with low probability over an extended period ensures that the attack is undetectable, but the impact is significant.

\subsection{Limitations and Future Work}
While our work presents a robust framework for TIRA attacks that expose the adversarial vulnerabilities in fairness tools and post-hoc explainability, several avenues for future research are still unexplored.

\textbf{Expanding Attack Vectors:} Our study is effective on binary and continuous protected features. Future work should extend the applicability of such attacks to more complex datasets and explainability tools.

\textbf{Theoretical Guarantees:} Our work empirically demonstrates the TIRA attacks' efficacy, but a deeper theoretical investigation is required. Proving the asymptotic properties of such attacks as well as formalizing their relationship to SHAP’s estimation algorithms would provide a much-needed theoretical foundation.

\textbf{Developing Robust Defenses:} The most pressing next step is the development of robust defenses. Future work should focus on designing auditing frameworks that can detect inconsistencies between a model’s outputs and its explanations.

\end{document}